\documentclass[runningheads]{llncs}
\usepackage{lmodern}
\usepackage[T1]{fontenc}
\usepackage[utf8]{inputenc}
\usepackage{color}
\usepackage{amsmath}
\usepackage{graphicx}
\PassOptionsToPackage{normalem}{ulem}
\usepackage{ulem}

\makeatletter

\providecommand{\tabularnewline}{\\}

\usepackage{amsmath}  
\usepackage{amsfonts}
\usepackage{algpseudocode}

\newcommand{\indep}{\perp \!\!\! \perp}

\makeatother

\begin{document}
\newcommand{\sidenote}[1]{\marginpar{\small \emph{\color{Medium}#1}}}

\global\long\def\model{\text{\text{\emph{cm-BO}}}}%

\title{Amortized Conditional Independence Testing}
\author{\vspace{-0.3cm}}
\author{Bao Duong, Nu Hoang\thanks{Corresponding author}, Thin Nguyen}
\institute{\vspace{-0.2cm}Applied Artificial Intelligence Institute (A2I2),
Deakin University, Australia\\ \email{\{duongng, nu.hoang, thin.nguyen\}@deadkin.edu.au}}

\maketitle
%

%
\begin{abstract}
\vspace{-0.7cm}

 Testing for the conditional independence structure in data is a
fundamental and critical task in statistics and machine learning,
which finds natural applications in causal discovery--a highly relevant
problem to many scientific disciplines. Existing methods seek to design
explicit test statistics that quantify the degree of conditional dependence,
which is highly challenging yet cannot capture nor utilize prior knowledge
in a data-driven manner. In this study, an entirely new approach is
introduced, where we instead propose to \textit{amortize conditional
independence testing} and devise \textbf{ACID} (\textbf{\uline{A}}mortized
\textbf{\uline{C}}onditional \textbf{\uline{I}}n\textbf{\uline{D}}ependence
test)--a novel transformer-based neural network architecture that
\textit{learns to test for conditional independence}. \textbf{ACID}
can be trained on synthetic data in a supervised learning fashion,
and the learned model can then be applied to any dataset of similar
natures or adapted to new domains by fine-tuning with a negligible
computational cost. Our extensive empirical evaluations on both synthetic
and real data reveal that \textbf{ACID} consistently achieves state-of-the-art
performance against existing baselines under multiple metrics, and
is able to generalize robustly to unseen sample sizes, dimensionalities,
as well as non-linearities with a remarkably low inference time.


\vspace{-0.3cm}
\end{abstract}

\section{Introduction\vspace{-0.3cm}}

Conditional independence testing is a cornerstone of statistical and
machine learning research, with critical applications in causal discovery
\cite{jaber2020causal,versteeg2022local,mooij2020constraint}, which
has broad implications across scientific fields. The Conditional Independence
(CI) testing problem is formulated as follows: given empirical data
of random variables $X$, $Y$, and $Z$, we seek to answer whether
or not $X$ and $Y$ are statistically independent after all the influences
from $Z$ to $X$ and $Y$ have been accounted for. More formally,
the problem can be described in the hypothesis testing framework with:
\vspace{-0.5cm}

\begin{align}
\mathcal{H}_{0} & :X\indep Y\mid Z\quad\text{and}\quad\mathcal{H}_{1}:X\not\indep Y\mid Z
\end{align}

\noindent where $\mathcal{H}_{0}$ and $\mathcal{H}_{1}$ are termed
as the \textit{Null hypothesis} and the \textit{Alternative hypothesis},
respectively. Following \cite{bellot2019conditional,Duong_Nguyen_2022Conditional},
we refer to $X$ and $Y$ as the \textit{target variables} and $Z$
as the \textit{conditioning variable}.

Ideally, the conditional independence $X\indep Y\mid Z$ can be checked
using the probability density/mass function $p\left(\cdot\right)$
of the underlying distribution, i.e., $\mbox{\ensuremath{X\indep Y\mid Z}}$
if and only if $p\left(x,y\mid z\right)=p\left(x\mid z\right)p\left(y\mid z\right)$
for all $x,y,z$ in the supports. This condition is \textit{intractable}
to verify for continuous and high-dimensional empirical data, thus
many methods must resort to discretization \cite{diakonikolas2016new,warren2021wasserstein},
which suffers from the curse of dimensionality. This intrinsic challenge
of CI testing and its significant impacts motivate us to tackle the
problem under the most challenging setting, where all variables are
continuous and may be multi-dimensional.

Traditionally, there is a common strategy to design a CI testing method,
which formulates a so-called \textit{test statistics} to quantify
the conditional dependence between the target variables given the
conditioning variable. Then, the distribution of the test statistics
under $\mathcal{H}_{0}$, i.e., the \textit{Null distribution}, is
derived. Next, given a test statistics value $t$ inferred from input
data, the Null distribution is used to compute the probability that
any dataset coming from $\mathcal{H}_{0}$ has a test statistics more
``extreme'' than $t$, which is called the test's \textit{$p$-value}.
Finally, if the $p$-value is less than a pre-specified significance
level $\alpha$ then $\text{\ensuremath{\mathcal{H}_{0}}}$ is rejected,
otherwise we fail to reject $\mathcal{H}_{0}$. While being mathematically
justified, this strategy entails two challenges. First, the test statistics
must well reflect the degree of conditional dependence and should
be efficiently estimated from data, which is technically challenging
and may involves restrictive assumptions. For example, while partial
correlation is a simple measure of conditional dependence that can
be easily computed, it is only valid when data follows a jointly multivariate
Gaussian or similar distribution \cite{baba2004partial}; meanwhile,
conditional mutual information is a powerful measure, but it is intractable
to compute exactly, thus requires approximations \cite{Samo_2021Inductive,kubkowski2021gain,Duong_Nguyen_2023Diffeomorphic}.
Second, to estimate the $p$-value, the null distribution must be
easily derived from data, either analytically or empirically. As an
example, the partial correlation is not used as-is for CI testing
because its null distribution is complex even for multivariate Gaussian
data, so the Fisher's z-transformation was devised to transform it
to an approximately Gaussian variable \cite{hotelling1953new}. Moreover,
when no analytical transformation can be found to obtain a simple
null distribution, bootstrapping is adopted to simulate the null distribution
from empirical data \cite{runge2018conditional,mukherjee2020ccmi},
in which the user must trade-off between the computational expense
and accuracy via the number of bootstraps. \vspace{-0.5cm}

\paragraph{Present study.}

Given the obstacles discussed above in designing a high-quality test
statistic, this study offers an entirely different approach to test
for conditional independence, motivated by the transition of learning
algorithms from machine learning to deep learning. Particularly,
deep learning (DL) exhibits a huge technological leap from classical
machine learning, where features are \textit{hand-crafted,} to the
stage where features are \textit{learned} by effectively utilizing
data. This capability has fueled a surge in developing heuristic models
for diverse tasks, such as causal structure learning \cite{Lorch_etal_2022Amortized,Ke_etal_2023Learning}
and combinatorial optimization \cite{khalil2017learning,shrivastava2020glad}.
Building on this momentum, we propose to innovate CI testing by \textit{learning
to test for conditional independence}. In other words, unlike existing
methods that demand designing specific test statistics, we amortize
conditional independence testing via a neural architecture that analyzes
the \textit{entire dataset at once}. This network is trained in a
supervised manner to determine whether the null hypothesis $\mathcal{H}_{0}$
is rejected or not. Designing such a network is tricky, requiring
to be flexible enough to handle datasets with varying numbers of variables
and robust to the order of the data, across all both samples and variables.
 To this end, we leverage the transformer architecture \cite{Vaswani_etal_2017Attention},
a breakthrough invention in deep learning, to design a new architecture
for CI testing, coined \textbf{ACID} (Amortized Conditional Independence
testing), which can be trained end-to-end on simulated data, where
the label is available, to reach arbitrary levels of competence in
CI testing. Notably, the trained model demonstrates impressive performance
when tested with unseen datasets, especially generalizing to various
data characteristics like sample size, dimensionality, and non-linearity,
coupled with a near-zero inference expense, effectively amortizing
its training cost.

The main contributions in this study include: \vspace{-0.15cm}
\begin{enumerate}
\item We introduce a novel approach for conditional independence testing,
by amortizing the procedure with a neural network that can learn to
perform conditional independence tests directly from data. To the
best of our knowledge, this is the first study exploring this promising
territory, contributing a significantly new perspective to the literature
of the field.
\item We present \textbf{ACID}, a tailor-made neural architecture that can
\textbf{} can operate with variable-sized datasets and is invariant
to permutations in both the sample and variable dimensions, maximizing
its statistical efficacy.
\item We demonstrate the effectiveness of the proposed \textbf{ACID} architecture
in comparison with state-of-the-art CI tests on a wide range of experiments
over synthetic datasets and real data. 
\end{enumerate}

\vspace{-0.5cm}

\section{Proposed Method \vspace{-0.35cm}}

\begin{figure*}[t]
\centering{}\includegraphics[width=1\textwidth]{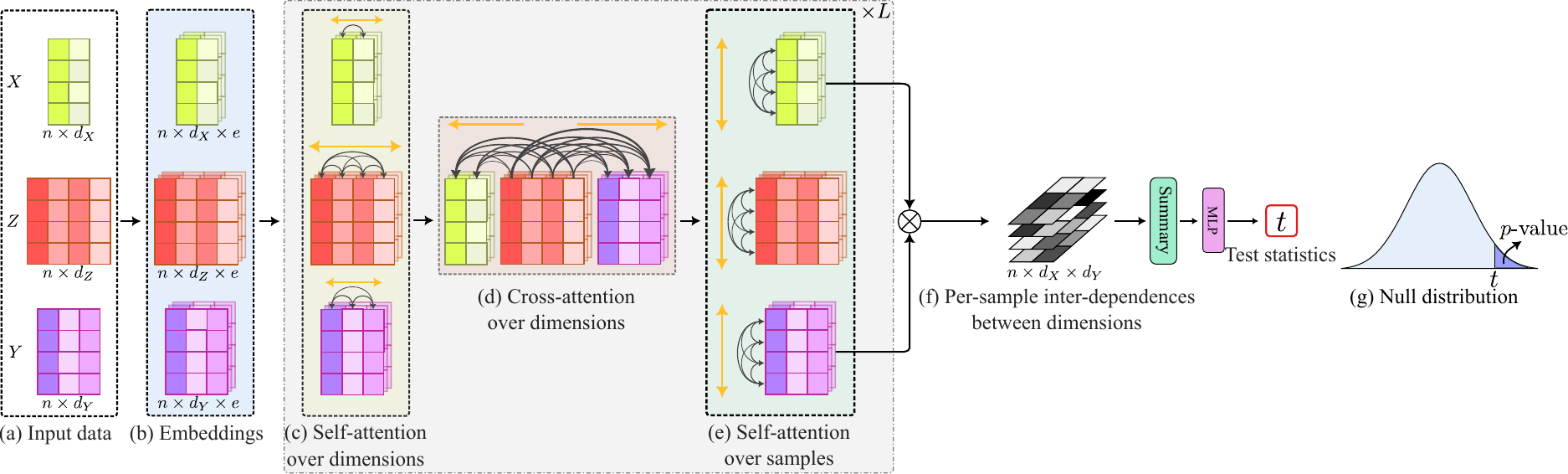}\caption{Overview of the proposed \textbf{ACID} architecture.  }\label{fig:illustration}
\vspace{-0.7cm}
\end{figure*}

In this section we describe in details the proposed \textbf{ACID}
architecture for CI testing, depicted in Figure~\ref{fig:illustration}.
\vspace{-0.5cm}

\subsection{A supervised learning approach for CI testing}

We denote an i.i.d. dataset as \scalebox{0.8}{ $\mathcal{D}=\left\{ \left(x^{\left(i\right)}\in\mathbb{R}^{d_{X}},y^{\left(i\right)}\in\mathbb{R}^{d_{Y}},z^{\left(i\right)}\in\mathbb{R}^{d_{Z}}\right)\right\} _{i=1}^{n}$}
and its label $\mathcal{G}\in\left\{ 0,1\right\} $ corresponds to
two classes $\mathcal{H}_{0}$ and $\mathcal{H}_{1}$, respectively.
The innovation in this study is seeded from the switch of perspective,
from seeing $\mathcal{D}$ as a mere dataset to seeing $\mathcal{D}$
as a \textit{random variable} in the context of CI testing. Thus,
there must exist a data generating distribution $p\left(\mathcal{D},\mathcal{G}\right)$
between datasets and labels. Once both are observed, we can lean to
predict a dataset's label by estimating the conditional distribution
of the label given the dataset $p\left(\mathcal{G}\mid\mathcal{D}\right)$.
Towards learning this conditional, we model it as $p_{\theta}\left(\mathcal{G}\mid\mathcal{D}\right):=\mathrm{Bernouli}\left(\mathcal{G};\mathcal{F}\left(\mathcal{D},\theta\right)\right)$,
where $\mathcal{F}\left(\cdot,\theta\right)$ is a neural network
parametrized by $\theta$ that maps a dataset to the parameter (typically
the real-valued logit) of the Bernoulli distribution. Then, the neural
network can be trained via minimizing the binary cross entropy (BCE)
loss commonly employed in binary classification:

\vspace{-0.5cm}

\begin{align}
\theta^{\ast} & :=\underset{\theta}{\arg\min}\ \mathcal{L}\left(\theta\right),\;\text{where}
\end{align}
 \vspace{-1cm}

{\small
\begin{equation}
\mathcal{L}\left(\theta\right):=-\mathbb{\mathbb{E}}_{p\left(\mathcal{D},\mathcal{G}\right)}\left[\mathcal{G}\ln p_{\theta}\left(\mathcal{G}\mid\mathcal{D}\right)+\left(1-\mathcal{G}\right)\ln p_{\theta}\left(\bar{\mathcal{G}}\mid\mathcal{D}\right)\right]\label{eq:loss}
\end{equation}
}{\small\par}

To accommodate the fact that $\mathcal{G}$ is usually not observed,
we leverage synthetic data where an unlimited amount of labeled data
is available, allowing us to train the neural network $\mathcal{F}$
to the fullest extent possible.

The remaining challenge for us now is to design a sophisticate neural
network that can take as input a \textit{whole dataset}, which is
significantly different from typical applications where a neural network
only takes a \textit{sample} as input. Not only so, the CI problem
also requires the neural network to be agnostic of the dataset's size
and arrangement, since shuffling data samples or dimensions must not
change the CI outcome. To achieve this goal, the Transformer architecture
\cite{Vaswani_etal_2017Attention} is the most well-fit device for
two reasons. First, all of its learnable parameters does not depend
on the dataset's size, hence the learned model can be applied to arbitrarily
sized datasets, enabling us to design a size-independent representation
of a whole dataset. The second reason involves the inherent \textit{permutation-invariance}
property of attention mechanisms, allowing us to encode the dataset
in a way that shuffling the dimensions or samples does not change
the prediction.

\vspace{-0.5cm}

\subsection{Model architecture}

\subsubsection{Multi-head attention}

Attention mechanism \cite{Vaswani_etal_2017Attention} was introduced
to construct a representation for one data sequence by attending and
aggregating information from another data sequence. In traditional
applications, attention mechanisms are adopted for modeling sequentially
ordered data, such as sentence or temporal data in general. However,
in our application the order must not be accounted for due to the
i.i.d. nature of the samples and the irrelevance between the dimensions
arrangement w.r.t. the CI label. Following \cite{Lorch_etal_2022Amortized,Ke_etal_2023Learning},
we detail below how the attention is customized for the CI testing
problem. 

Assuming we have two data matrices $A\in\mathbb{R}^{n\times d_{A}}$
and $B\in\mathbb{R}^{m\times d_{B}}$ where we wish to exploit information
from $B$ to produce a new representation for $A$. One of the most
common attention mechanisms--the scaled dot-product attention \cite{Vaswani_etal_2017Attention}--computes
the attention weights, i.e., how much to attend to each position,
by comparing the queries $Q=AW^{Q}\in\mathbb{R}^{n\times e}$ with
the keys $K=BW^{K}\in\mathbb{R}^{m\times e}$ using their dot-product,
scaled by the squared root of the embedding size $e$, then the updated
representation is the aggregation of the values $V=BW^{V}\in\mathbb{R}^{m\times e}$
weighted by the softmax-activated attention weights:

\begin{equation}
\mathrm{attn}\left(A,B\right):=\mathrm{softmax}\left(\frac{QK^{\top}}{\sqrt{e}}\right)V\in\mathbb{R}^{n\times e}
\end{equation}

\noindent where $W^{Q}\in\mathbb{R}^{d_{A}\times e}$, $W^{K}\in\mathbb{R}^{d_{B}\times e}$,
and $W^{V}\in\mathbb{R}^{d_{B}\times e}$ are learnable projection
matrices that linearly map the inputs $A$ and $B$ to the query,
key, and value spaces, respectively.\textit{ Self-attention} is a
special form of attention where $A=B$. Conversely, \textit{cross-attention}
is the attention specifically for two different entities $A\neq B$.\textit{
Multi-head attention} is an extension of the ordinal attention mechanism,
which attends to different aspects of the reference sequence, empowering
a richer representation for the target sequence. To achieve this,
multi-head attention performs a series of $h$ independent attentions,
then concatenate and mix them altogether:

{\small
\begin{equation}
\mathrm{mha}_{h}\left(A,B\right):=\mathrm{concat}\left(H_{1},\ldots,H_{h}\right)W^{H}\in\mathbb{R}^{n\times e}
\end{equation}
}{\small\par}

\noindent where $H_{i}:=\mathrm{attn}\left(AW_{i}^{A},BW_{i}^{B}\right)$
is an attention head with the inputs linearly projected to another
space via the learnable weight matrices $W_{i}^{A}\in\mathbb{R}^{d_{A}\times e}$
and $W_{i}^{B}\in\mathbb{R}^{d_{B}\times e}$, and $W^{H}\in\mathbb{R}^{\left(e\cdot h\right)\times e}$
is another learnable weight matrix that mixes the outputs of different
attention heads.

In addition, following the best practices in Transformer-based models
\cite{Vaswani_etal_2017Attention,Xiong_2020Layer,Kossen_etal_2021Self},
we first add a skip connection \cite{He_etal_2016Deep}, then a Dropout
\cite{Srivastava_etal_2014Dropout}, followed by Layer Normalization
\cite{Ba_etal_2016Layer} to create an intermediate representation:

{\small
\begin{equation}
\mathrm{IR}\left(A,B\right):=\mathrm{LN}\left(AW^{Q}+\mathrm{Dropout}\left(\mathrm{mha}_{h}\left(A,B\right)\right)\right)
\end{equation}
}{\small\par}

Then, we apply a position-wise feed-forward network (FFN) \cite{Vaswani_etal_2017Attention},
followed by Dropout and another skip connection, and obtain the full
multi-head attention layer for our architecture:

\begin{equation}
\mathrm{MHAttn}\left(A,B\right):=AW^{Q}+\mathrm{Dropout}\left(\mathrm{FFN}\left(\mathrm{IR}\left(A,B\right)\right)\right)\label{eq:mhatnn}
\end{equation}

When multi-head self-attention of $A$ is used, we simply denote the
output as $\mathrm{MHAttn}\left(A\right)$.

\vspace{-0.5cm}

\subsubsection{Whole-dataset Encoder}

Given a dataset \scalebox{0.8}{$\mathcal{D}=\left\{ \left(x^{\left(i\right)},y^{\left(i\right)},z^{\left(i\right)}\right)\right\} _{i=1}^{n}$},
we first stack the samples of each random variables to create data
matrices $X\in\mathbb{R}^{n\times d_{X}}$, $Y\in\mathbb{R}^{n\times d_{Y}}$,
and $Z\in\mathbb{R}^{n\times d_{Z}}$. The goal of the dataset encoder
is to discover the intrinsic statistical properties of the given dataset
to infer what makes it conditionally independent of dependent. With
the help of attention mechanisms, our data encoder exploits the intra-dimensional,
inter-dimensional, as well as intra-sample relationships presented
in the dataset. These are performed in multiple layers, where each
layer is expected to produce a finer representation of the dataset.
In our implementation, $L=4$ layers of three types of attentions
are used. Additionally, it is important to note that $X$ and $Y$
share the same role in the CI condition, because $X\indep Y\mid Z$
is equivalent with $Y\indep X\mid Z$. To preserve this symmetry in
our architecture, every operator applied on $X$ is applied identically
on $Y$ with the same set of parameters. Hence, for brevity, in the
remaining part of this section we omit the descriptions of the operators
on $Y$, which are similar to $X$.

\vspace{-0.3cm}

\paragraph{Data entry embedding}

Before we can encode the dataset, we first need to embed each entry
in the data matrices from a scalar to a vector of size $e$. The embeddings
are obtained by passing each individual data entry to a shared position-wise
feed-forward network (FFN). More particularly, we use two FFNs namely
$\mathrm{Emb}_{XY}\left(\cdot\right):\mathbb{R}\rightarrow\mathbb{R}^{e}$
shared between $X$ and $Y$, and $\mathrm{Emb}_{Z}\left(\cdot\right):\mathbb{R}\rightarrow\mathbb{R}^{e}$
for $Z$ solely. This process results in the initial representations
of $X$, $Y$, and $Z$, which we denote as
\begin{align}
\mathcal{X}^{\left(0\right)} & :=\mathrm{Emb}_{XY}\left(X\right)\in\mathbb{R}^{n\times d_{X}\times e}\\
\mathcal{Z}^{\left(0\right)} & :=\mathrm{Emb}_{Z}\left(Z\right)\in\mathbb{R}^{n\times d_{Z}\times e}
\end{align}

\paragraph{Self-attention over dimensions (SoD)}

We are now in a position to encode the dataset. We proceed by applying
(multi-head) self-attention (Equation~\ref{eq:mhatnn}) to each variable
to capture their intra-dimensional relationships. Particularly, for
the $l$-th layer ($l=1\ldots L$), we perform the following independently
and identically for every row $i=1\ldots n$:{\small
\begin{align}
\mathrm{SoD}_{XY}^{\left(l\right)}\left(\mathcal{X}_{i}^{\left(l-1\right)}\right) & :=\mathcal{\mathrm{MHAttn}}\left(\mathcal{X}_{i}^{\left(l-1\right)}\right)\\
\mathrm{SoD}_{Z}^{\left(l\right)}\left(\mathcal{Z}_{i}^{\left(l-1\right)}\right) & :=\mathcal{\mathrm{MHAttn}}\left(\mathcal{Z}_{i}^{\left(l-1\right)}\right)
\end{align}
}{\small\par}

\noindent where the $\mathrm{SoD}$ modules are shared across all
samples, therefore they are unaffected by the arrangement of the samples.

\paragraph{Cross-attention over dimensions (CoD) from $Z$ to $X$ and $Y$}

We next fuse the information of $Z$ into $X$ and $Y$ with the end
goal being the representations of $X$ and $Y$ that contain minimal
information about $Z$. While this may sound contradictory, it is
inspired from the observation that $X\indep Y\mid Z$ if and only
if after removing all information of $Z$ from $X$ and $Y$, the
remaining of $X$ and $Y$ are marginally independent. This is mathematically
supported by \cite{Duong_Nguyen_2022Conditional}, where it is proved
that if there exists invertible functions $f_{X}$ and $f_{Y}$ such
that $\left(f_{X}\left(X,Z\right),f_{Y}\left(Y,Z\right)\right)\indep Z$,
i.e., they remove all information of $Z$ from $X$ and $Y$, then
$X\indep Y\mid Z\Leftrightarrow f_{X}\left(X,Z\right)\indep f_{Y}\left(Y,Z\right)$.
Our CoD module mimics this feature of these functions. Specifically,
at layer $l$ we distribute the information of $Z$ into $X$ and
$Y$ by applying the following identically for each row $i=1\ldots n$
of $X$ and $Y$:{\small
\begin{align}
\mathrm{CoD}^{\left(l\right)}\left(\mathcal{X}_{i}^{\left(l-1\right)},\mathcal{Z}_{i}^{\left(l-1\right)}\right) & :=\mathrm{MHAttn}\left(\mathcal{X}_{i}^{\left(l-1\right)},\mathcal{Z}_{i}^{\left(l-1\right)}\right)
\end{align}
}{\small\par}

\paragraph{Self-attention over samples (SoS)}

While the first two attention modules exploit the dependence mechanism
across data dimensions, our last attention module detects the distributional
characteristics of the dataset by attending over the datapoints, as
suggested in \cite{Kossen_etal_2021Self}. To be more specific, the
$l$-th layer of SoS is defined identically for each \textit{column}
$j$ of the data matrices as:{\small
\begin{align}
\mathrm{SoS}_{XY}^{\left(l\right)}\left(\mathcal{X}_{:,j}^{\left(l-1\right)}\right) & :=\mathrm{MHAttn}\left(\mathcal{X}_{:,j}^{\left(l-1\right)}\right)\\
\mathrm{SoS}_{Z}^{\left(l\right)}\left(\mathcal{Z}_{:,j}^{\left(l-1\right)}\right) & :=\mathrm{MHAttn}\left(\mathcal{Z}_{:,j}^{\left(l-1\right)}\right)
\end{align}
}{\small\par}

In summary, the final output of each attention layer is the sum of
the three kinds of attention above with a skip connection, followed
by an FFN for increased capacity:{\footnotesize
\begin{align}
\mathcal{X}^{\left(l\right)} & :=\mathrm{FFN}_{XY}^{\left(l\right)}\left(\mathcal{X}^{\left(l-1\right)}+\mathrm{SoD}_{XY}\left(\cdot\right)+\mathrm{CoD}\left(\cdot\right)+\mathrm{SoS}_{XY}\left(\cdot\right)\right)\\
\mathcal{Z}^{\left(l\right)} & :=\mathrm{FFN}_{Z}^{\left(l\right)}\left(\mathcal{Z}^{\left(l-1\right)}+\mathrm{SoD}_{Z}\left(\cdot\right)+\mathrm{SoS}_{Z}\left(\cdot\right)\right)
\end{align}
}{\footnotesize\par}

\noindent where we omit the input arguments to avoid notational clutter.

\paragraph{Encoder summary}

After a series of attention layers, as argued above, we expect the
dependence between $\mathcal{X}^{\left(L\right)}$ and $\mathcal{Y}^{\left(L\right)}$
can expose the conditional dependence of the original variables, which
motivates us to measure their dependence. Towards this end, we begin
by extending the embedding dimension by $h$ times, which correspond
to $h$ inter-dependence heads that focus on different aspects of
the data, inspired by multi-head attentions. This is done via a single
FFN $\mathbb{R}^{e}\rightarrow\mathbb{R}^{h\times e}$ as follows:

{\small
\begin{align}
\tilde{\mathcal{X}} & :=\mathrm{FFN}\left(\mathcal{X}^{\left(L\right)}\right)\in\mathbb{R}^{n\times d_{X}\times h\times e}\\
\tilde{\mathcal{Y}} & :=\mathrm{FFN}\left(\mathcal{Y}^{\left(L\right)}\right)\in\mathbb{R}^{n\times d_{Y}\times h\times e}
\end{align}
}{\small\par}

Then, we measure the sample-level inter-dependence magnitudes between
the dimensions of $\tilde{\mathcal{X}}$ and $\tilde{\mathcal{Y}}$
using the squared dot-product of the embeddings:

{\small
\begin{equation}
\mathcal{S}:=\left(\tilde{\mathcal{X}}\otimes\tilde{\mathcal{Y}}\right)^{2}\in\mathbb{R}^{n\times d_{X}\times d_{Y}\times h}
\end{equation}
}{\small\par}

Next, we average this over the sample axis and take the maximum magnitude
over every pairs of dimensions as the final $h$-dimensional vector
representation of the dataset, where the $r$-th value corresponds
to the $r$-th inter-dependence head:

{\small
\begin{equation}
\mathcal{R}_{r}:=\max_{i,j}\frac{1}{n}\sum_{o=1}^{n}\mathcal{S}_{o,i,j,r}
\end{equation}
}{\small\par}

Since $\sum$ is invariant to samples permutations and $\max$ is
invariant to dimensions permutations, $\mathcal{R}$ is permutation-invariant
to both samples and dimensions of $\mathcal{D}$. Additionally, in
our experiments, the number of attention heads and inter-dependence
heads are the same and equal to $h=8$.

\vspace{-0.5cm}

\subsubsection{Classification}

The final representation vector of the dataset is used as input for
a Multiple-layer Perceptron (MLP) $\mathcal{T}:\mathbb{R}^{h}\rightarrow\mathbb{R}$
that outputs the logit of the conditional distribution $p_{\theta}\left(\mathcal{G}\mid\mathcal{D}\right)$.
In other words, $p_{\theta}\left(\mathcal{G}\mid\mathcal{D}\right)$
is a Bernoulli distribution with mean equals to $\mathrm{sigmoid}\left(\mathcal{T}\left(\mathcal{R}\left(\mathcal{D}\right)\right)\right)$.
Plugging this into Equation~\ref{eq:loss} we obtain a fully differentiable
system that is end-to-end trainable with any gradient-based optimization
method. In our implementation, the popular momentum-based Adam optimizer
\cite{Kingma_Ba_2016Adam} is employed for this purpose.

\vspace{-0.5cm}

\subsection{Test statistic and Null distribution}

Once trained, the predicted logit is used as our test statistics since
they are predictive of the CI label by design. As for the null distribution,
we collect the predicted logits of training datasets with label $\mathcal{G}=0$
to construct an empirical null distribution. Our experiments (Section~\ref{subsec:Training})
show that this distribution is simple with a normal-like shape. Therefore,
we fit it with a slightly more flexible distribution, i.e., the skewed
normal distribution \cite{O_Leonard_1976Bayes}, and use it to calculate
the $p$-value for all datasets. Here, higher values of the test statistics
are considered more extreme since datasets from $\mathcal{H}_{0}$
should have the logits, i.e., the test statistics, as low as possible.
Hence, the $p$-value in our framework is the extreme region on the
right tail of the null distribution:

\begin{equation}
p\text{-value}\left(\mathcal{D}\right):=1-\Phi\left(\mathcal{T}\left(\mathcal{R}\left(\mathcal{D}\right)\right)\right)
\end{equation}

\noindent where $\Phi$ is the cumulative distribution function of
the fitted skewed normal distribution.

\vspace{-0.35cm}

\section{Experiments \vspace{-0.25cm}}

\begin{figure}[t]
\centering
\begin{minipage}[c]{0.3\columnwidth}%
\caption{ The learning process of \textbf{ACID}. Validation dataset is not
needed since no dataset is seen twice during training. Shaded areas
depict standard deviations. }\label{fig:Training-summary}
\end{minipage}%
\begin{minipage}[t]{0.02\columnwidth}%
\textcolor{white}{blank}%
\end{minipage}%
\begin{minipage}[c]{0.7\columnwidth}%
\includegraphics[scale=0.5]{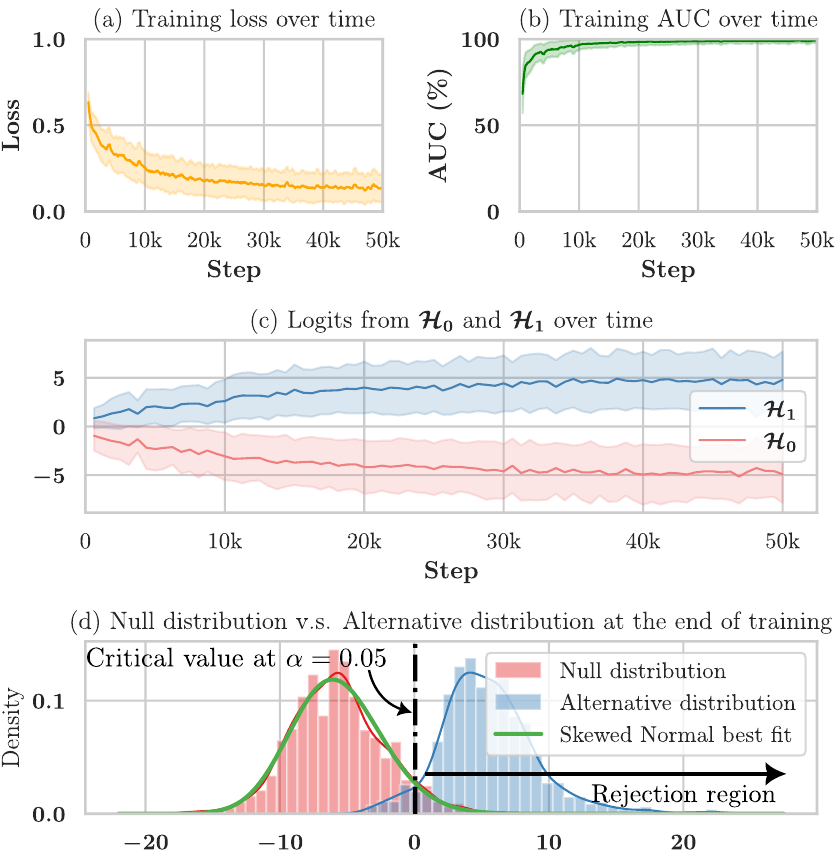}%
\end{minipage}
\end{figure}

\begin{figure}[t]
\begin{centering}
\begin{tabular}{c}
\includegraphics[width=0.9\textwidth]{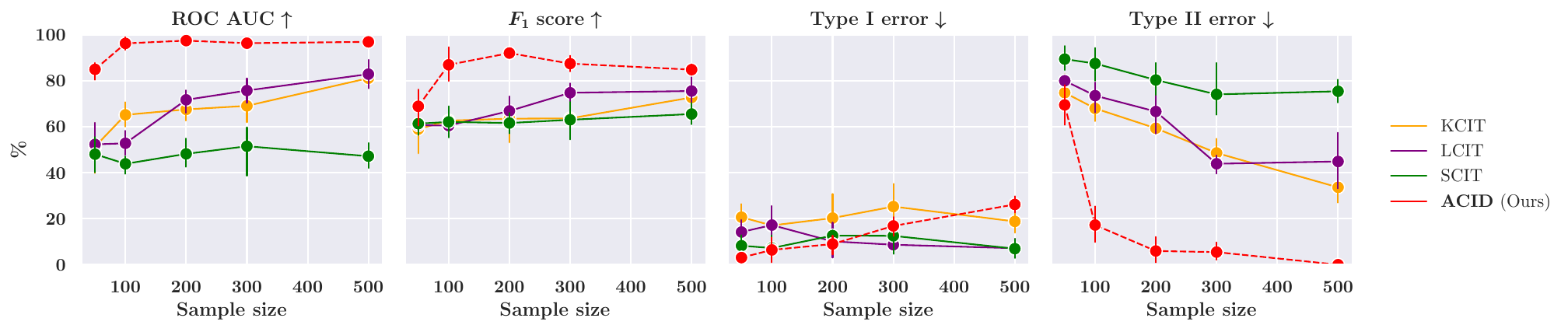}\tabularnewline
(a) Conditional Independence Testing performance w.r.t. different
sample sizes\tabularnewline
\includegraphics[width=1\textwidth]{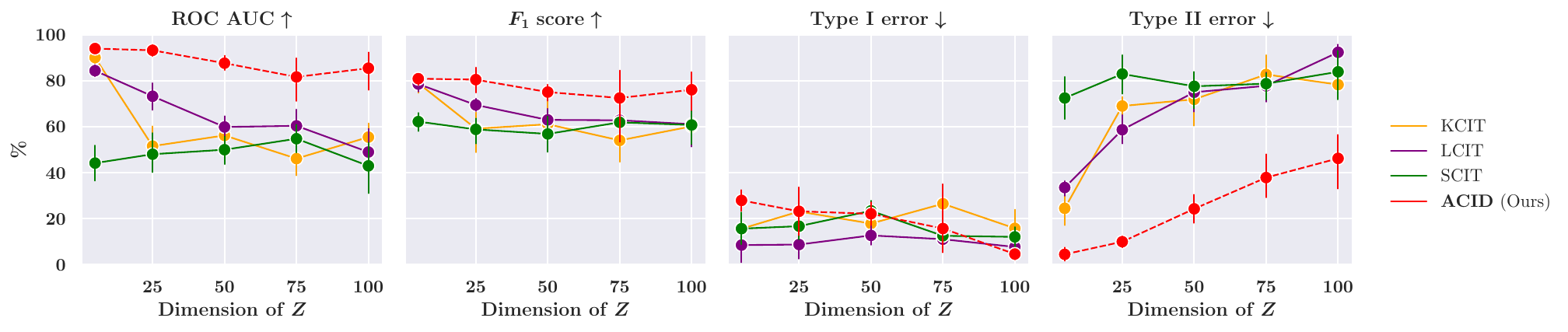}\tabularnewline
(b) Conditional Independence Testing performance w.r.t. different
dimensionalities of $Z$\tabularnewline
\end{tabular}
\par\end{centering}
\centering{}\caption{Conditional Independence Testing performance on synthetic data. The
evaluation metrics are AUC, $F_{1}$ score (higher is better), Type
I and Type II errors (lower is better). Error bars are 95\% confidence
intervals.\vspace{-0.15cm}}\label{fig:synthetic}
\end{figure}

\begin{figure}[t]
\centering
\begin{minipage}[c]{0.3\columnwidth}%
\caption{Conditional Independence Testing performance in Out-of-distribution
settings. The performance metrics are AUC, $F_{1}$ score (higher
is better), Type I and Type II errors (lower is better). Error bars
are 95\% confidence intervals.}\label{fig:ood}
\end{minipage}%
\noindent\begin{minipage}[t]{0.01\columnwidth}%
\textcolor{white}{blank}%
\end{minipage}%
\begin{minipage}[c]{0.69\columnwidth}%
\begin{center}
\includegraphics[scale=0.15]{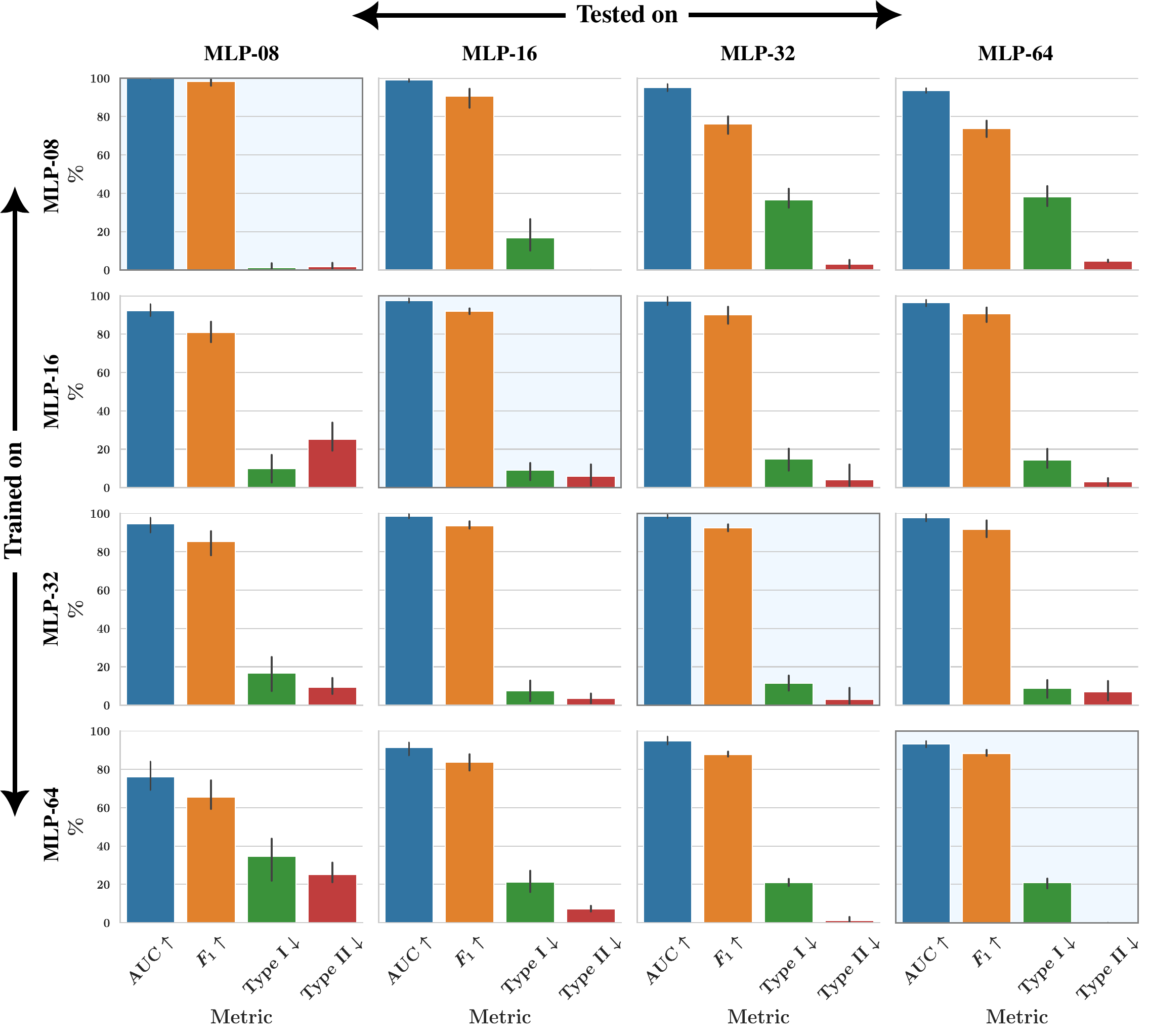}
\par\end{center}%
\end{minipage}
\end{figure}

\begin{figure}[t]
\centering
\begin{minipage}[c]{0.4\columnwidth}%
\caption{Conditional Independence Testing performance on Real data (the Sachs
dataset \cite{Sachs_etall_05Causal}). Time is measured on an Apple
M1 CPU with 8 GB of RAM. }\label{fig:real}
\end{minipage}%
\begin{minipage}[t]{0.005\columnwidth}%
\textcolor{white}{blank}%
\end{minipage}%
\begin{minipage}[c]{0.7\columnwidth}%
\begin{center}
\includegraphics[width=0.7\columnwidth]{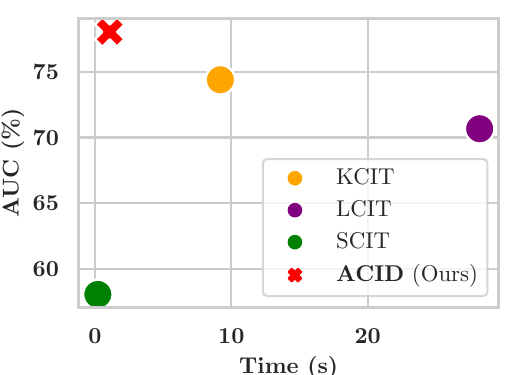}
\par\end{center}%
\end{minipage}
\end{figure}

\subsection{Experimental Setting}

\subsubsection{Methods and evaluation criterion}

We compare the CI testing performance of the proposed \textbf{ACID}
test in comparison with both popular and recent state-of-the-arts.
Specifically, we consider KCIT \cite{zhang2012kernel}--a well known
and widely adopted kernel-based method, SCIT \cite{zhang2022residual}--a
recent development from regression-based methods, and LCIT \cite{Duong_Nguyen_2022Conditional}--a
recent DL approach based on representation learning. Regarding evaluation
metrics, we employ Area Under the Receiver Operating Characteristic
Curve (ROC AUC, or AUC), $F_{1}$ score, Type I, and Type II errors.

\vspace{-0.5cm}

\subsubsection{Synthetic data}

For the synthetic data generation process to be as generic as possible
so models trained on this scheme can have robust generalizability,
we generate data following six data models of conditional (in)dependences
based on three canonical types of graphical $d$-separation \cite{Pearl2009}.
To incorporate functional mechanisms of varying non-linearity, we
employ MLPs as the function class for the links between the variables.
We denote MLP-$k$ as the class of MLPs with one hidden layer of size
$k$ and we control the level of non-linearity via the choice of $k$.
In all comparative experiments, we generate 200 i.i.d. datasets
from the simulated distribution with 100 datasets belonging to class
$\mathcal{H}_{0}$ and 100 datasets belonging to $\mathcal{H}_{1}$.
To account for uncertainty, we randomly divide them into five separate
folds for estimating the metrics' variation. Moreover, to make sure
no test data is seen during training, we use separate random seed
ranges for the training and test data. 

\vspace{-0.5cm}

\subsection{Results on Synthetic Data}

\subsubsection{Training of ACID}\label{subsec:Training}

We first train \textbf{ACID} on datasets with $n\in\left\{ 50,200\right\} $,
$d_{X}=d_{Y}=1$, $d_{Z}\in\left\{ 5,10,20\right\} $, and $k=16$.
Batches of fresh synthetic datasets are continuously generated for
training \textbf{ACID} so that a dataset is very rarely seen twice,
maximizing the model's generalizability on the training domain and
avoid overfitting. This model is trained strictly on small-scale datasets
to demonstrate the generalizability to larger-scale data (Section~\ref{subsec:Generalization-to-Sample}
and Section~\ref{subsec:Generalization-to-Dimensionalities}). Figure~\ref{fig:Training-summary}
presents the training summary. The training process converges after
50,000 steps at a loss of $0.1$. However, the model clearly separates
the $\mathcal{H}_{0}$ and $\mathcal{H}_{1}$ logits and achieves
an AUC of nearly 100\% after just 10,000 steps, which takes only about
50 minutes on a system with an Nvidia V100 32GB GPU. This confirms
that the \textbf{ACID} architecture is indeed able to test for conditional
independence given sufficient training data, even if the datasets
possess a limited number of samples.\vspace{-0.5cm}

\subsubsection{Generalization to Sample sizes}\label{subsec:Generalization-to-Sample}

We next test \textbf{ACID}'s ability to generalize to datasets with
different sample sizes than what it was trained on. Specifically,
we fix $d_{X}=d_{Y}=1$, $d_{Z}=10$ and compare the CI testing performance
of \textbf{ACID} against state-of-the-arts in MLP-16 with sample sizes
varying from 5 to 500. The result is reported in Figure~\ref{fig:synthetic}(a),
which shows that \textbf{ACID} is far more effective than its baseline
counterparts for all sample sizes. More particularly, from 100 samples
to 500 samples, our method always achieves the highest AUC scores
of nearly 100\%, vanishing Type II errors, highest $F_{1}$ scores
of at least 80\%, and Type I errors competitive with baseline methods.
At the extreme case of only 5 samples per dataset, \textbf{ACID} still
achieves the highest AUC of over 80\%, while other competitors struggle
at 50\%. Additionally, even though with just 5 samples our Type II
error is up to 70\% due to the lack of information, the situation
is quickly mitigated as the sample size increases to 100, where the
Type II error of \textbf{ACID} reaches below 20\% but other state-of-the-arts
still struggle at over 70\%.\vspace{-0.5cm}

\subsubsection{Generalization to Dimensionalities}\label{subsec:Generalization-to-Dimensionalities}

To evaluate the generalizability of the proposed \textbf{ACID} architecture
over different dimensionalities of the conditioning variable $Z$,
we fix $n=300$, $d_{X}=d_{Y}=1$, and vary $d_{Z}$ from 5 to 100
dimensions. The result depicted in Figure~\ref{fig:synthetic}(b)
demonstrates the remarkable extendability of \textbf{ACID}, which
was trained on data of only 20 dimensions but still consistently surpasses
all baselines in 100 dimensions, as evidenced by the highest AUC and
$F_{1}$ scores, as well as lowest Type II errors in all cases, while
Type I error is comparable to other competitors. \vspace{-0.5cm}

\subsubsection{Out-of-distribution Generalization to Non-linearity}\label{subsec:Generalization-to-Non-linearity}

In the following, we examine \textbf{ACID}'s ability to generalize
beyond the training distribution. Particularly, we consider four different
data domains with exponentially increasing difficulties, namely MLP-08,
MLP-16, MLP-32, and MLP-64. Each environment is used to train an \textbf{ACID}
model, which is subsequently cross-validated on the other domains.
In Figure~\ref{fig:ood} we display the out-of-distribution generalization
results. The hardness of the considering environments is reflected
in the in-distribution numbers, where the easiest environment MLP-08
can be learned to the fullest extent, with near 100\% AUC and $F_{1}$
scores, as well as vanishing Type I and Type II errors. Meanwhile,
models trained on the most challenging environment MLP-64 can only
reach 90\% in AUC and $F_{1}$, with 20\% of Type I error.  \vspace{-0.5cm}

\subsection{Performance and Scalability on Real Data}

To validate the effectiveness of the proposed \textbf{ACID }test,
we demonstrate evaluate it against state-of-the-arts in CI testing
on the Sachs dataset \cite{Sachs_etall_05Causal}, a common benchmarking
real dataset. We adapt our model to the real data by fine-tuning
a model in Section~\ref{subsec:Training} for only 500 steps on really
small down-sampled datasets of merely 50 samples. The process takes
no more than three minutes on a system with an Nvidia V100 32GB GPU.
Then, in Figure~\ref{fig:real} we compare the performance of all
methods along with their runtime. The time is measured on an Apple
M1 CPU with 8 GB of RAM. It can be seen that our method achieves the
best performance both in terms of AUC and inference time. With under
one second of inference time for each dataset of 853 samples and $d_{Z}=3$,
we achieve 78\% of AUC, while the second-best method, which is KCIT,
can only achieve 74\%, with an average inference time of 10 seconds
per dataset. Meanwhile, SCIT is the fastest method but obtains the
lowest AUC score among all. \vspace{-0.5cm}

\section{Conclusions \vspace{-0.25cm}}

In this study we propose a novel supervised learning approach to conditional
independence testing called \textbf{ACID}, which amortizes the conditional
independence testing process via a neural network that can learn to
test for conditional independence. The performance of our \textbf{ACID}
architecture is demonstrated via a wide range of experiments on both
synthetic and real datasets, which show that \textbf{ACID} can generalize
well beyond the trained data and consistently outperform existing
state-of-the-arts. Our architecture has potential to be extended furthermore
to handle more challenging data scenarios, such as mixed-type or missing
value, which will be the subject of future development.

\paragraph{\vspace{-1.0cm}}

\bibliographystyle{splncs04}
\bibliography{ref}

\end{document}